\documentclass{article}

\PassOptionsToPackage{numbers, compress}{natbib}

\usepackage{tech2025}

\usepackage[utf8]{inputenc} 
\usepackage[T1]{fontenc}    
\usepackage{url}            
\usepackage{booktabs}       
\usepackage{amsfonts}       
\usepackage{amsmath}        
\usepackage{amssymb}
\usepackage{nicefrac}       
\usepackage{microtype}      
\usepackage{xcolor}         
\usepackage{multirow}
\usepackage{appendix}
\usepackage{algorithmicx} 
\usepackage{wrapfig}
\usepackage{subcaption}
\usepackage[utf8]{inputenc} 
\usepackage[T1]{fontenc}    
\usepackage{url}            
\usepackage{booktabs}       
\usepackage{amsfonts}       
\usepackage{amsmath}        
\usepackage{nicefrac}       
\usepackage{microtype}      
\usepackage{xcolor}         
\usepackage{multirow}
\usepackage{appendix}
\usepackage{tabularx}
\usepackage{makecell}
\usepackage{wrapfig}
\usepackage{subcaption}
\usepackage{bbding}
\usepackage{todonotes}
\usepackage{fvextra}
\usepackage[pro]{fontawesome5}

\usepackage{algorithm}
\usepackage{algpseudocode}
\usepackage{graphicx}  
\usepackage{bbm}
\usepackage{float}
\usepackage{xcolor,colortbl}
\usepackage{amsmath}
\usepackage{braket}
\usepackage{amssymb}
\usepackage{mathtools}
\usepackage{amsthm}
\usepackage{amsmath,amsfonts,bm}
\usepackage{bigstrut,bigdelim}
\usepackage{paralist}
\usepackage{diagbox}
\usepackage{wrapfig}
\usepackage{verbatim}
\usepackage{framed}
\usepackage[most]{tcolorbox}

\usepackage{subcaption}
\usepackage{multicol}
\usepackage{multirow}
\usepackage{amssymb}
\usepackage[tableposition=top]{caption}
\usepackage{hyperref}       
\usepackage{cleveref}
\usepackage{acronym}
\usepackage{xspace}
\usepackage{pifont}
\usepackage{lipsum}
\usepackage{textcomp,scalerel}
\usepackage{enumitem}
\usepackage[misc]{ifsym}
\usepackage{appendix}
\usepackage{titletoc}
\usepackage{lipsum}
\usepackage{tabularx}
\usepackage{makecell}

\makeatletter
\DeclareRobustCommand\onedot{\futurelet\@let@token\@onedot}
\def\@onedot{\ifx\@let@token.\else.\null\fi\xspace}

\makeatother

\makeatletter
\renewcommand{\paragraph}{%
  \@startsection{paragraph}{4}%
  {\z@}{0ex \@plus 0ex \@minus 0ex}{-1em}%
  {\hskip\parindent\normalfont\normalsize\bfseries}%
}
\makeatother

\makeatletter
\newcommand{\thickhline}{%
    \noalign {\ifnum 0=`}\fi \hrule height 1pt
    \futurelet \reserved@a \@xhline
}
\makeatother

\theoremstyle{plain}

\theoremstyle{definition}

\theoremstyle{remark}

\acrodef{psrm}[PSRM]{Perfect Sparse Reward Model}
\acrodef{tta}[TTIA]{Test-Time Instance-Level Adaptation}
\acrodef{cot}[CoT]{Chain-of-Thought}

\def\vx{\mathbf{x}}
\def\vc{\mathbf{c}}
\def\vz{\mathbf{z}}

\usepackage{natbib}

\definecolor{bigaired}{RGB}{156, 0, 0}
\definecolor{uclablue}{RGB}{39, 116, 174}

\definecolor{darkred}{RGB}{200, 0, 0}
\definecolor{darkblue}{RGB}{0, 0, 200}
\definecolor{blue}{RGB}{0, 0, 250}

\definecolor{light}{RGB}{225, 250, 250}
\definecolor{lightgray}{RGB}{0.9, 0.9, 0.9}
\definecolor{lightred}{RGB}{250, 200, 200}
\definecolor{lightblue}{RGB}{210, 220, 250}

\definecolor{doderblue}{RGB}{30, 144, 255}
\definecolor{select}{RGB}{222, 235, 247}
\definecolor{unselect}{RGB}{247, 207, 206}

\definecolor{lightgrey}{RGB}{247, 247, 247}

\hypersetup{colorlinks=true, citecolor=uclablue, linkcolor=uclablue, urlcolor=bigaired}

\DeclareUnicodeCharacter{0394}{\ensuremath{\Delta}}
\DeclareUnicodeCharacter{03C0}{\ensuremath{\pi}}

\newcommand{\model}{\textsc{GradCuit}\xspace}

\usepackage{natbib}

\begin{document}

\makeatletter
\def\icmldate#1{\gdef\@icmldate{#1}}
\icmldate{\today}
\makeatother

\newenvironment{bigaiabstract}{
  \begin{tcolorbox}[
    colback=lightgrey,
    colframe=white,
    boxrule=0pt,
    arc=10pt,
    left=16pt,
    right=16pt,
    top=12pt,
    bottom=12pt,
    width=\textwidth,
    enlarge left by=0mm,
    before skip=10pt,
    after skip=10pt
  ]
  \normalsize
}{
  \end{tcolorbox}
}

\makeatletter
\fancypagestyle{fancytitlepage}{
  \fancyhead{}  
  \lhead{\includegraphics[height=1.2cm]{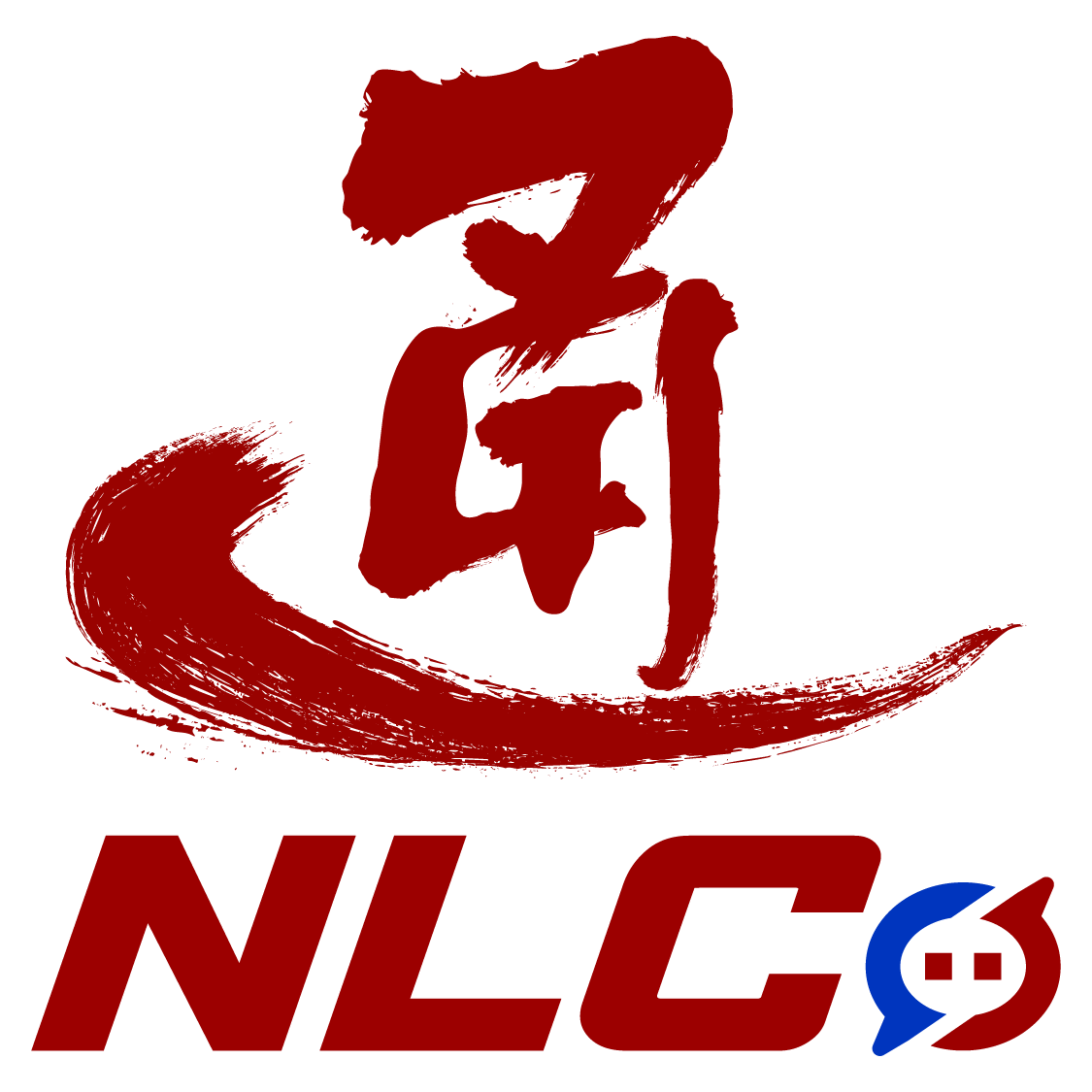}
  \includegraphics[height=1.2cm]{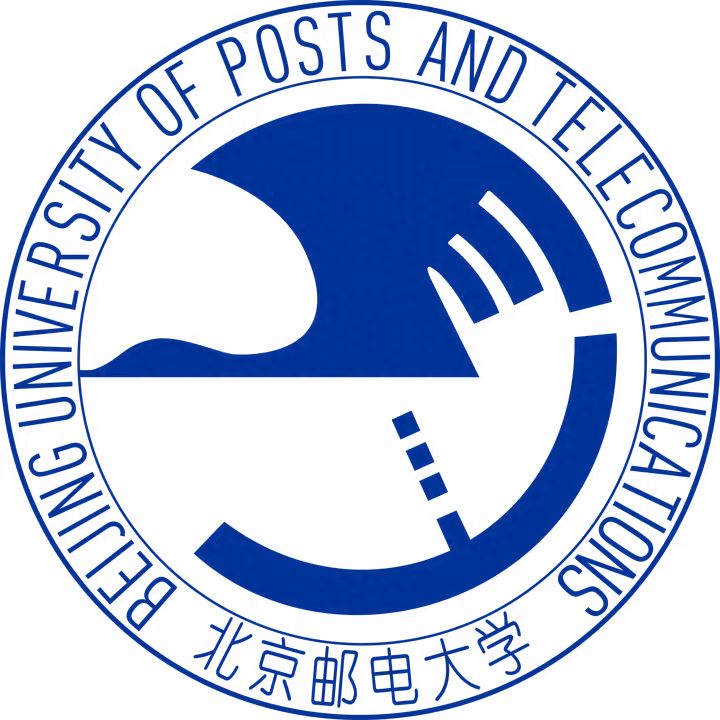}
  \includegraphics[height=1.2cm]{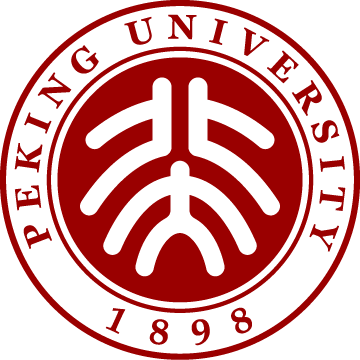}
}
  \rhead{\it \@icmldate}
  \cfoot{}
}
\makeatother


\icmltitle{GradCuit: Credit-Assigned Gradient Flow Enables Robust and Interpretable Test-Time Latent Reasoning}

\begin{icmlauthorlist}
Zhaoxin Yu$^{\,1\,*}$, Qi Shen$^{\,1,2\,*}$, Hengli Li$^{\,1,3\,*\,\dagger}$,  Zhaowei Zhang$^{\,3}$, Song-Chun Zhu$^{\,1}$, Chi Zhang$^{\,1,3\,}$\textsuperscript{\Letter} and Zilong Zheng$^{\,1\,}$\textsuperscript{\Letter}
\end{icmlauthorlist}

$^{1\,}$NLCo Lab, Beijing Institute for General Artificial Intelligence  \\
$^{2\,}$School of Artificial Intelligence, Beijing University of Posts and Telecommunications \\
\quad $^{3\,}$School of Artificial Intelligence for Science, Peking University

{\small\texttt{zhaoxin.yu010@gmail.com, shenqi@bupt.edu.cn,lihengli@stu.pku.edu.cn, chizhang.cz@pku.edu.cn, zlzheng@bigai.ai}}

\icmlcorrespondingauthor{Hengli Li}{lihengli.cs@stu.pku.edu.cn}
\icmlcorrespondingauthor{Chi Zhang}{chizhang.cz@pku.edu.cn}
\icmlcorrespondingauthor{Zilong Zheng}{zlzheng@bigai.ai}

\printNotice{\textsuperscript{*}~Equal Contributions.
  \textsuperscript{\textdagger}~Project lead.
  \textsuperscript{\Letter}~Equal Supervision.} 

\vskip .3in

\begin{bigaiabstract}

Optimization-based latent reasoning improves large language model outputs by optimizing instance-specific continuous states at test time while keeping model parameters frozen. Existing methods, however, typically connect these states to the reasoning trajectory through decoded tokens, making sequence-level credit assignment indirect and obscuring how latent updates shape subsequent reasoning. We introduce \model\ (\emph{gradient through circuit}), which inserts optimizable latent states at a selected Transformer layer between the hidden representations of the prompt and the generated continuation. Causal self-attention provides every continuation-token log-probability with a differentiable path to every preceding latent state through the remaining Transformer blocks, enabling reward-weighted gradients from the entire continuation to be assigned directly to the latents. Across five instruction-tuned backbones, three reasoning benchmarks, and two answer formats, \model achieves an average accuracy of 64.5\%, outperforming chain-of-thought prompting by 6.6 percentage points and the strongest competing method by 2.4 points. \model also demonstrates greater \textbf{robustness}: across seven learning-rate settings, it consistently outperforms LatentSeek while reducing the standard deviation of accuracy from 1.53 to 0.82, and even its random-walk variant remains competitive with LatentSeek. For \textbf{interpretability}, token-level gradient attribution reveals that latent influence concentrates on reasoning-connector tokens, while layer analysis identifies early-to-middle Transformer layers as the most effective optimization space. By directly optimizing internal reasoning from outcome feedback, \model opens a new axis of robust and interpretable test-time scaling, where LLMs adapt how they reason rather than merely regenerate, sample, or rerank outputs.

\begin{center}  
\setlength{\tabcolsep}{5pt}
\begin{tabular}{ccl}
   \faGithub  & \textbf{Code} & \url{https://github.com/Yuzhaoxin946/GradCuit} \\
   \faGlobe & \textbf{Project} &\url{https://yuzhaoxin946.github.io/GradCuit}
\end{tabular}

\end{center}

\end{bigaiabstract}
\vskip .3in

\section{Introduction}
\label{sec:introduction}

A growing line of work explores augmenting large language models (LLMs) with \emph{latent reasoning} \citep{hao2024training,li2026seekdarkreasoningtesttime,deng2024explicit}, where continuous latent variables serve as intermediate states beyond explicit chain-of-thought tokens \citep{wei2022chain}. Recent reasoning-as-optimization methods, including LatentSeek \citep{li2026seekdarkreasoningtesttime}, LTPO \citep{ye2025thinkingflytesttimereasoning}, and MILR \citep{mi2025milrimprovingmultimodalimage}, optimize instance-specific latents at test time to improve generation quality without updating model parameters. However, since they rely on decoded tokens as the interface between latent variables and the reasoning path, optimization is restricted to policy gradients of token-level objectives, which can only reach the latents by backpropagating through the decoding process.
 This designation creates a fundamental \textbf{credit-assignment} challenge: it remains difficult to determine how latent variables contribute to subsequent reasoning tokens and the final answer.

To be specific, this challenge manifests in two aspects:
(1) \textbf{Indirect optimization}: the decoding process introduces an information bottleneck between latent variables and their downstream effects, causing optimization signals to become indirect and entangled with intermediate token representations;
(2) \textbf{Opaque latent dynamics}: the discrete generation pathway obscures how individual latent variables influence subsequent predictions, making it difficult to interpret a particular latent update.

\begin{figure}[t]
    \centering
    \includegraphics[width=\textwidth]{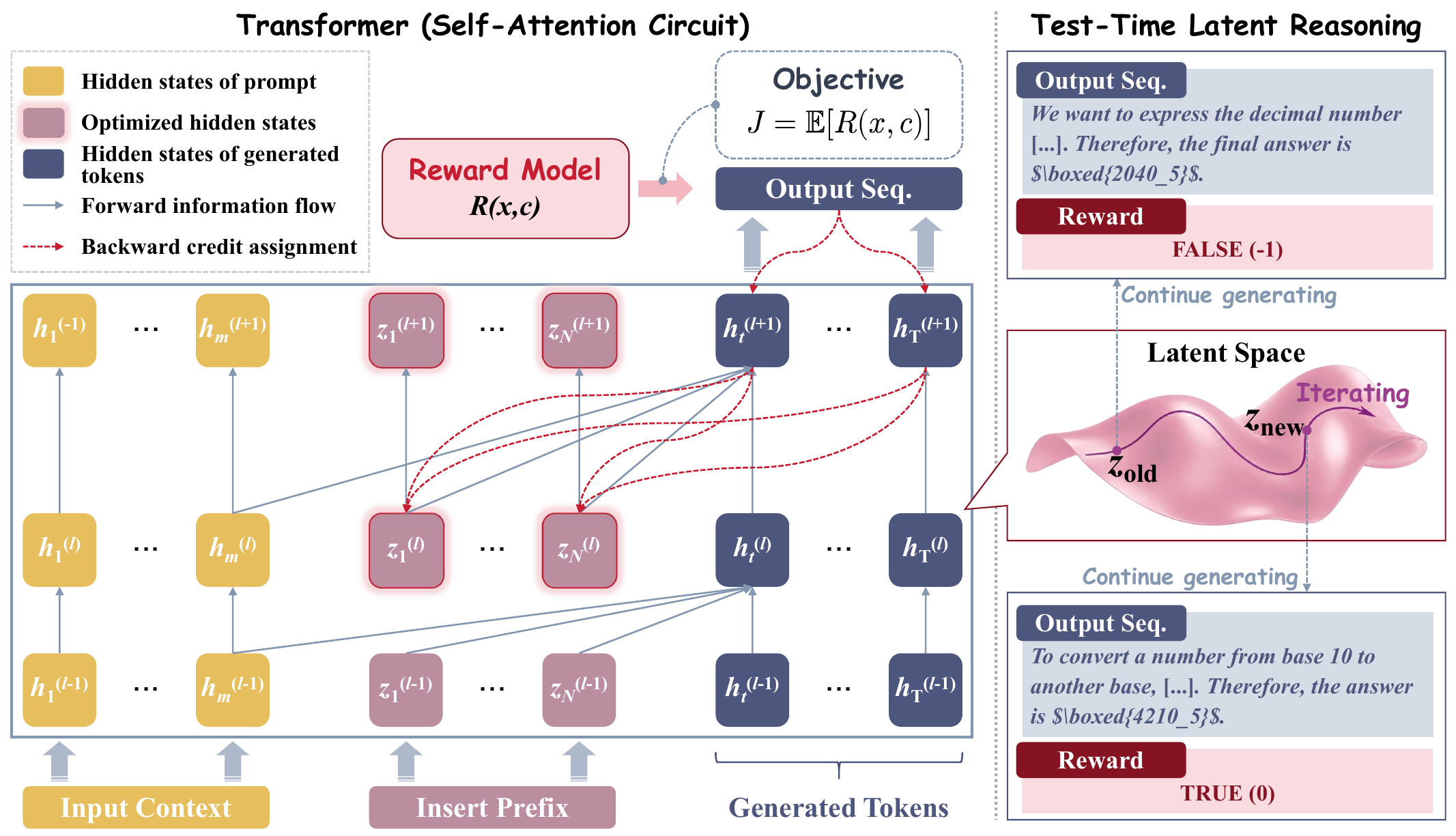}
    \caption{
        The \model framework. The Transformer's self-attention mechanism
        functions as a gradient-routing circuit, enabling direct backward
        credit assignment from generated tokens to the latents along
        attention pathways. The red feedback loop visualizes the policy
        gradient update
        ($z \leftarrow z + \nabla_z \mathcal{J}$),
        used to iteratively refine the latents during inference without
        updating the base model parameters.
    }
    \label{fig:gradcuit}
\end{figure}

We propose \model, which takes a step toward addressing this credit-assignment challenge by leveraging the Transformer's self-attention mechanism as a \emph{computational circuit} for latent reasoning. Our design choice is to introduce latents directly into the model's \emph{intermediate hidden space}, allowing the self-attention circuit to serve as both the forward pathway for computation and the backward pathway for gradient propagation during optimization. This designation is related in spirit to the \emph{Transformer Circuits} viewpoint \citep{elhage2021mathematical}: the attention graph defines a compositional circuit of interactions, hence the name \model\ (\emph{gradient through circuit}). \model enables \textbf{direct optimization} of the latents through self-attention, while simultaneously offering \textbf{a lens into the latent dynamics} \footnote{Concurrent work by Anthropic, J-lens \citep{gurnee2026verbalizable}, explores a related idea. However, whereas their focus is on explainability, our goal is to improve latent reasoning.} via the Jacobian of the subsequent tokens with respect to the latents.

Extensive experiments across five backbones and three reasoning benchmarks demonstrate the strong and consistent performance of \model. Averaged across all backbones, benchmarks, and answer formats, \model improves accuracy over the standard CoT baseline by 6.6 percentage points. It also outperforms all four enhanced reasoning baselines, achieving an overall average accuracy of 64.5\%, compared with 62.1\% for the strongest competing method. \model further exhibits greater \textbf{robustness}: across seven learning-rate settings, it consistently outperforms LatentSeek while reducing the standard deviation in accuracy from 1.53 to 0.82. Even when its policy-gradient updates are replaced by a random walk, \model remains competitive with LatentSeek \citep{li2026seekdarkreasoningtesttime}, demonstrating the robustness of its latent space to the choice of optimization direction. Finally, this circuit-like gradient pathway enables direct inspection of latent dynamics, revealing the important role of reasoning-connector tokens and thereby improving the \textbf{interpretability} of latent reasoning.

\paragraph{Contributions.}

\begin{itemize}
\item \textbf{Circuit-like latent reasoning through self-attention.}
We introduce \model, which inserts a small set of learnable latent states at an intermediate Transformer layer, enabling self-attention to provide a direct computational pathway for both forward token--latent interaction and backward credit assignment.

\item \textbf{Improved robustness through direct credit assignment.}
By propagating reward-weighted gradients directly from generated tokens to the latent states, \model achieves greater optimization robustness than previous latent-reasoning methods.

\item \textbf{Effective and interpretable latent dynamics.}
Across five backbones and three reasoning benchmarks, \model improves accuracy over CoT by 6.6 percentage points and outperforms the strongest enhanced reasoning baseline by 2.4 points on average. Its circuit-like gradient pathway also makes latent dynamics more interpretable, revealing the important role of reasoning-connector tokens in latent reasoning.
\end{itemize}

\section{Circuit-like Gradient Flow for Latent Reasoning}
\label{sec:methodology}
In this section, we present the \model algorithm.

\subsection{Preliminaries: Output-Side Latent Optimization}
Let $\pi$ be an autoregressive language model and $R$ a reward model. For an input problem $\vc$, let $\vx=(x_1,x_2,\ldots,x_T)$ denote a length-$T$ token sequence. As a representative output-side formulation, LatentSeek~\citep{li2026seekdarkreasoningtesttime} introduces latent variables $\vz=(z_1,z_2,\ldots,z_N)$, where each $z_t$ lies in the latent space associated with $x_t$ and $N \le T$. Its ideal test-time objective is to maximize the expected reward:
\begin{equation}
\label{eq:latent_reasoning_objective}
\vz^*=\arg\max_{\vz}\ \mathbb{E}_{\vx \sim \pi(\vx \mid \vz,\vc)}\bigl[R(\vx,\vc)\bigr].
\end{equation}

The conditional generation distribution $\pi(\vx \mid \vz,\vc)$ is factorized as
\begin{equation}
\label{eq:latentseek_factorization}
\pi(\vx \mid \vz,\vc)
= \underbrace{\prod_{t=1}^{N}\pi_\theta(x_t \mid z_t)}_{\text{Latents }\to\text{ tokens}}
\ \prod_{t=N+1}^{T}\pi(x_t \mid \vx_{<t},\vc),
\end{equation}
where $\theta$ parameterizes the language-model head (LM head).

Conceptually, LatentSeek first decodes the latent variables $\vz$ into the first $N$ tokens, and then performs standard autoregressive generation conditioned on this decoded prefix. After producing the full sequence $\vx$ and evaluating its reward $R(\vx,\vc)$, LatentSeek updates the latents by backpropagating gradients through the logits associated with these decoded tokens. Concretely, the policy-gradient term for the $i$-th latent takes the form
\begin{equation}
\label{eq: reward weighted latent objective}
    \mathbb{E}_{\vx \sim \pi(\vx \mid \vz, \vc)}[R(\vx, \vc) \nabla_{z_i}\log \pi(x_i \mid z_i)].
\end{equation}

\subsection{Gradient Flow through Transformer Circuits}

To precisely attribute the influence of individual latent variables and avoid the information loss introduced by decoding continuous latents into discrete tokens, we propose a new method, \model, for solving \Cref{eq:latent_reasoning_objective}. Rather than defining the latent space at the transformer's \emph{output}, as in \citet{li2026seekdarkreasoningtesttime}, we define it within an \emph{intermediate} hidden-state space. This design directly incorporates the pre-trained self-attention mechanism into the latent optimization process.

Consider an $M$-layer transformer decoder, and let the output space of its $l$-th layer serve as the latent optimization space. To predict the $t$-th token $x_t$, we first pass the prompt $\vc$ and the previously generated tokens $\vx_{<t}$ through the first $l$ transformer layers, obtaining their corresponding hidden representations, $h^{(l)}_{\vc}$ and $h^{(l)}_{\vx_{<t}}$, respectively. We then insert the optimizable latent variables $\vz^{(l)}$ between these representations to form the concatenated sequence
\[
\bigl[h^{(l)}_{\vc},\, \vz^{(l)},\, h^{(l)}_{\vx_{<t}}\bigr].
\]

The next-token distribution is obtained by passing this concatenated sequence through the remaining Transformer layers and the language-model head:
\begin{equation}
\label{eq:calculating_next_token}
\begin{aligned}
\pi(x_t \mid \vx_{<t}, \vz^{(l)}, \vc)
&=
\operatorname{LM\_Head}\!\Bigl(\operatorname{Transformer}^{l+1:M}\!\bigl(
[h^{(l)}_{\vc},\, \vz^{(l)},\, h^{(l)}_{\vx_{<t}}]
\bigr)
\Bigr).
\end{aligned}
\end{equation}

This construction induces\textbf{a new autoregressive factorization} that replaces \Cref{eq:latentseek_factorization}:
\begin{equation}
\label{eq:new_factorization}
\pi(\vx \mid \vz^{(l)}, \vc)
=
\prod_{t=1}^{T}
\pi(x_t \mid \vx_{<t}, \vz^{(l)}, \vc).
\end{equation}

Because the \textbf{latent variables participate in self-attention alongside the token representations}, each generated token can attend to all preceding latent positions. Consequently, gradient signals from every generated token can propagate directly to every latent variable through the remaining self-attention layers. This token-to-latent gradient flow is characterized as
\[
\nabla_{z_i^{(l)}}\,
\pi(x_t \mid \vx_{<t}, \vz^{(l)}, \vc),
\]
which measures the sensitivity of the probability of the $t$-th token to the $i$-th latent variable. We accordingly perform the following policy-gradient-style update, as in LatentSeek~\citep{li2026seekdarkreasoningtesttime}:
\begin{equation}
\label{eq:gradcuit_policy_gradient}
\vz^{(l)}
\leftarrow
\vz^{(l)}
+
\eta \nabla_{\vz^{(l)}}\mathcal{J}(\vz^{(l)}),
\end{equation}
where $\eta$ denotes the latent optimization step size. The gradient associated with each latent variable aggregates contributions from all generated token positions:
\begin{equation}
\label{eq:token_attributed_gradient}
\begin{aligned}
\nabla_{z_i^{(l)}}\mathcal{J}
&=
\sum_{t=1}^{T}
\mathbb{E}_{\vx \sim \pi(\vx \mid \vz^{(l)}, \vc)}
\Big[R(\vx, \vc)\ \times \nabla_{z_i^{(l)}}
\log \pi(x_t \mid \vx_{<t}, \vz^{(l)}, \vc)
\Big].
\end{aligned}
\end{equation}

In summary, \textbf{gradient signals flow from the generated discrete tokens back to each latent variable through the transformer's self-attention connectivity}. Motivated by this circuit-like pathway for gradient propagation, we refer to the resulting latent reasoning method as \model.

\section{Experiments}
In this section, we present a comprehensive empirical evaluation of \model.
\subsection{Experimental Setup}

We evaluate \model against CoT, our one-pass Self-Reflection baseline, Self-Consistency, Self-Scored Best-of-$N$ (BoN), and \textsc{LatentSeek}~\citep{li2026seekdarkreasoningtesttime} to assess its effectiveness relative to representative explicit, sampling-based, and latent reasoning methods. Complete baseline protocols and prompts are provided in the Appendix.
Experiments are conducted using five instruction-tuned backbones: LLaMA-3.2-3B-Instruct,\footnote{\url{https://huggingface.co/meta-llama/Llama-3.2-3B-Instruct}} LLaMA-3.1-8B-Instruct,\footnote{\url{https://huggingface.co/meta-llama/Llama-3.1-8B-Instruct}} Qwen2.5-7B-Instruct,\footnote{\url{https://huggingface.co/Qwen/Qwen2.5-7B-Instruct}} Qwen2.5-14B-Instruct,\footnote{\url{https://huggingface.co/Qwen/Qwen2.5-14B-Instruct}} and Qwen3-4B-Instruct-2507.\footnote{\url{https://huggingface.co/Qwen/Qwen3-4B-Instruct-2507}}
We evaluate all methods on GPQA-Diamond~\citep{rein2024gpqa}, GSM8K~\citep{cobbe2021gsm8k}, and MATH-500~\citep{lightman2023let,hendrycksmath2021}, using both \texttt{\textbackslash boxed} and JSON answer formats. Full details are provided in the Appendix.





\begin{table*}[t]
\centering

\fontsize{8pt}{10.5pt}\selectfont
\begin{tabular}{lcccccccccccc}
\toprule
Backbone
& \multicolumn{2}{c}{LLaMA3.2-3B}
& \multicolumn{2}{c}{LLaMA3.1-8B}
& \multicolumn{2}{c}{Qwen2.5-7B}
& \multicolumn{2}{c}{Qwen2.5-14B}
& \multicolumn{2}{c}{Qwen3-4B}
& \multicolumn{2}{c}{Avg.} \\
\cmidrule(lr){1-1}
\cmidrule(lr){2-3}
\cmidrule(lr){4-5}
\cmidrule(lr){6-7}
\cmidrule(lr){8-9}
\cmidrule(lr){10-11}
\cmidrule(lr){12-13}
Prompt Type
& Boxed & JSON
& Boxed & JSON
& Boxed & JSON
& Boxed & JSON
& Boxed & JSON
& Boxed & JSON \\
\midrule

\multicolumn{13}{c}{GPQA-Diamond} \\
\midrule

CoT
& 17.7 & 18.7
& 20.7 & 22.2
& 31.3 & 31.8
& 40.4 & 42.4
& 40.9 & 40.4
& 30.2 & 31.1 \\

Self-Reflection
& 12.6 & \underline{25.3}
& 14.1 & 25.8
& 33.8 & 31.8
& 36.9 & \underline{45.5}
& \textbf{58.6} & \underline{54.6}
& 31.2 & 36.6 \\

Self-Consistency
& 23.2 & \underline{25.3}
& 25.8 & 28.8
& \underline{34.9} & 34.9
& 41.4 & 39.9
& 50.5 & \textbf{55.1}
& 35.2 & \underline{36.8} \\

Self-Scored BoN
& 18.2 & 24.2
& 15.7 & 26.3
& 33.3 & \textbf{37.9}
& 41.4 & \textbf{46.5}
& 46.5 & 48.0
& 31.0 & 36.6 \\

\textsc{LatentSeek}
& \underline{25.3} & 20.7
& \underline{27.8} & \underline{29.3}
& 33.3 & 32.3
& \textbf{47.5} & 42.9
& 50.0 & 51.0
& \underline{36.8} & 35.2 \\

\midrule

\textsc{GradCuit} (Ours)
& \textbf{30.3} & \textbf{27.8}
& \textbf{31.3} & \textbf{30.3}
& \textbf{38.4} & \underline{35.4}
& \underline{42.4} & 42.9
& \underline{52.5} & 49.5
& \textbf{39.0} & \textbf{37.2} \\

\midrule
\multicolumn{13}{c}{GSM8K} \\
\midrule

CoT
& 75.4 & 68.2
& 81.7 & 76.6
& 89.2 & 79.9
& 92.4 & 89.6
& 88.1 & 87.3
& 85.4 & 80.3 \\

Self-Reflection
& 67.9 & 58.5
& 70.9 & 67.3
& 89.2 & 79.4
& \underline{93.2} & 89.1
& \textbf{92.0} & 88.6
& 82.6 & 76.6 \\

Self-Consistency
& \underline{79.5} & \underline{69.7}
& \underline{84.0} & \underline{84.0}
& 88.2 & \underline{81.4}
& 92.2 & \underline{90.6}
& 89.2 & \underline{88.7}
& \underline{86.6} & \underline{82.9} \\

Self-Scored BoN
& 77.9 & 64.6
& 83.4 & 75.7
& 89.2 & 76.3
& 92.6 & 89.5
& 89.9 & 88.2
& \underline{86.6} & 78.9 \\

\textsc{LatentSeek}
& 77.8 & 68.8
& 82.6 & 78.2
& \underline{90.8} & 79.1
& 92.1 & 90.1
& 88.0 & 88.6
& 86.3 & 81.0 \\

\midrule

\textsc{GradCuit} (Ours)
& \textbf{82.5} & \textbf{75.7}
& \textbf{86.2} & \textbf{84.5}
& \textbf{91.8} & \textbf{82.1}
& \textbf{93.3} & \textbf{92.6}
& \underline{90.0} & \textbf{88.9}
& \textbf{88.8} & \textbf{84.8} \\

\midrule
\multicolumn{13}{c}{MATH-500} \\
\midrule

CoT
& 40.4 & 40.0
& 50.6 & 46.8
& 73.0 & 43.4
& 79.2 & 59.8
& 87.2 & 79.8
& 66.1 & 54.0 \\

Self-Reflection
& 35.4 & 38.4
& 38.4 & 40.2
& 74.6 & 47.0
& 76.4 & 62.0
& 84.4 & 81.6
& 61.8 & 53.8 \\

Self-Consistency
& 45.2 & \textbf{48.0}
& 46.4 & 50.8
& 73.8 & \underline{68.4}
& \underline{80.0} & \underline{68.4}
& 89.0 & \underline{84.6}
& 66.9 & \underline{64.0} \\

Self-Scored BoN
& 45.8 & 42.0
& 48.0 & 45.2
& 74.4 & 64.2
& 78.8 & 66.8
& \underline{90.0} & 83.4
& 67.4 & 60.3 \\

\textsc{LatentSeek}
& \underline{47.6} & 40.2
& \underline{56.8} & \underline{51.0}
& \underline{76.4} & 45.2
& 78.2 & 62.6
& 87.0 & 81.4
& \underline{69.2} & 56.1 \\

\midrule

\textsc{GradCuit} (Ours)
& \textbf{53.6} & \underline{47.4}
& \textbf{57.4} & \textbf{53.6}
& \textbf{77.2} & \textbf{68.6}
& \textbf{80.2} & \textbf{68.6}
& \textbf{91.8} & \textbf{86.8}
& \textbf{72.0} & \textbf{65.0} \\

\bottomrule
\end{tabular}
\caption{Accuracy comparison across different reasoning settings. 
\textsc{GradCuit} denotes our method.}
\label{tab:gradcuit_results}
\end{table*}

\subsection{Main Results}
\label{sec:main_experiments}

\textbf{Overall Effectiveness.}
Table~\ref{tab:gradcuit_results} summarizes the main results. Averaged over all 30 backbone--benchmark--format settings, \model achieves \textbf{64.5\%} accuracy, outperforming CoT by \textbf{6.6 points} and the strongest competing method by \textbf{2.4 points}. It attains the highest average accuracy under both answer formats on all three benchmarks and achieves the best individual result in \textbf{23 of 30 settings}. \model also consistently outperforms Self-Consistency and Self-Scored BoN in all six benchmark--format aggregates. Moreover, its average number of optimization iterations is lower than the number of sampled responses used by these methods in every setting; additional statistics are reported in the Appendix.

\textbf{Comparison with LatentSeek.}
\model improves over \textsc{LatentSeek} in every benchmark--format aggregate. Under the Boxed/JSON formats, the respective gains are \textbf{2.2/2.0 points} on GPQA-Diamond, \textbf{2.5/3.8 points} on GSM8K, and \textbf{2.8/8.9 points} on MATH-500. These consistent gains support the benefit of aggregating continuation-token gradients through the remaining Transformer computation, rather than restricting each latent's update to the log-probability of its associated decoded prefix token.

\begin{figure}[t]
\centering
\begin{minipage}[t]{0.48\columnwidth}
    \centering
    \includegraphics[width=\linewidth]
    {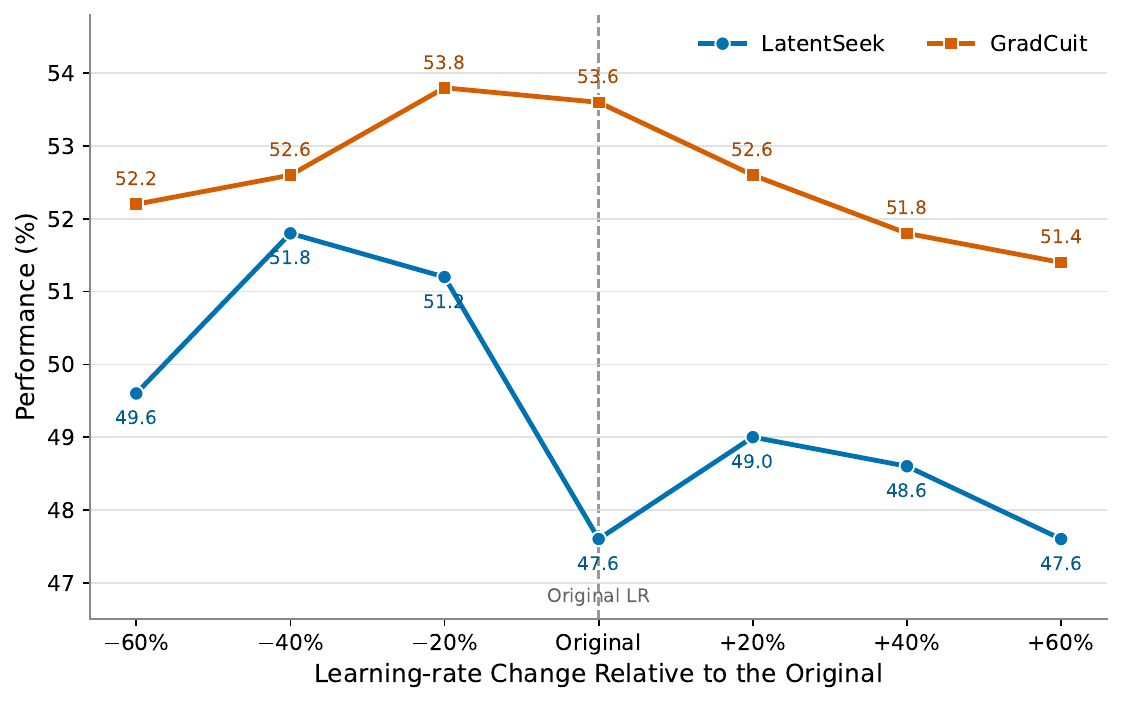}
    \caption{Learning-rate sensitivity of \textsc{LatentSeek} and \model on MATH-500 using LLaMA-3.2-3B-Instruct.}
    \label{fig:learning_rate_robustness}
\end{minipage}
\hfill
\begin{minipage}[t]{0.48\columnwidth}
    \centering
    \includegraphics[width=\linewidth]
    {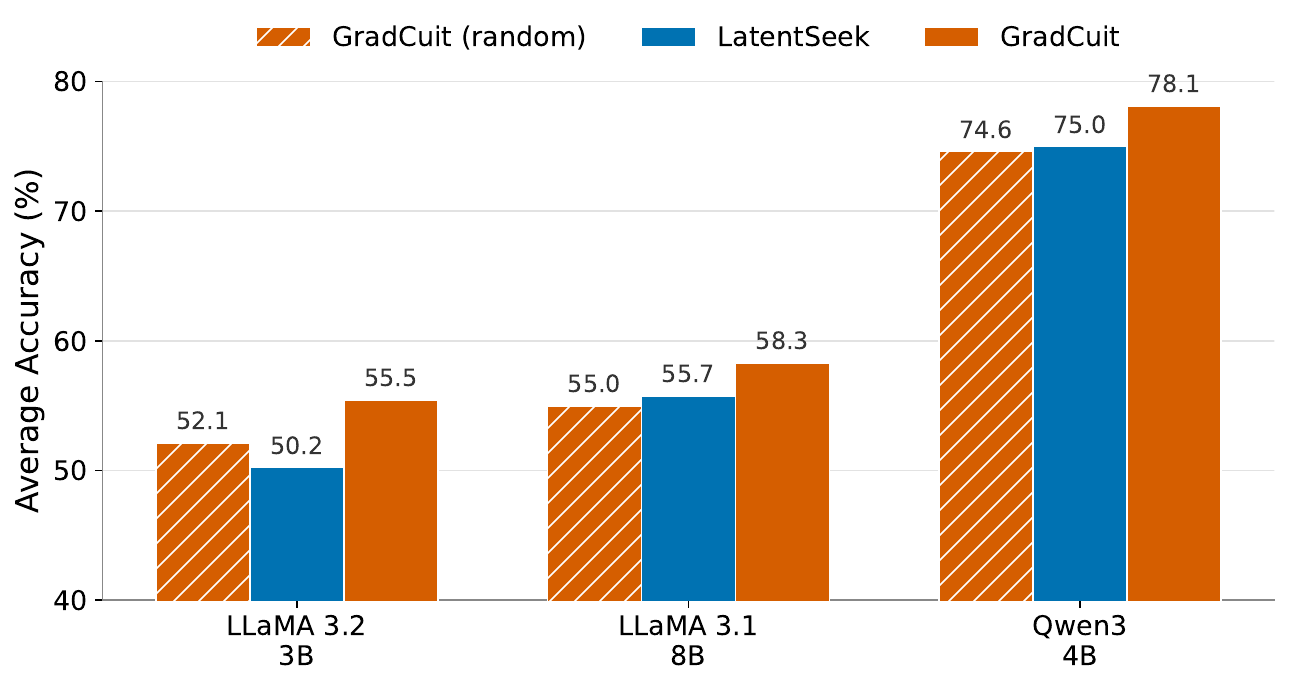}
    \caption{Average accuracy across three representative backbones. \model{} (random) uses randomly sampled update directions.}
    \label{fig:random_vs_latentseek}
\end{minipage}
\end{figure}

\subsection{Robustness Gain}
\label{sec:robustness_gain}

We examine the robustness of latent optimization from two complementary perspectives: sensitivity to the learning rate and dependence on reward guidance. For the learning-rate analysis, we evaluate \model and \textsc{LatentSeek} using LLaMA-3.2-3B-Instruct on MATH-500 with the Boxed answer format. We scale their respective base learning rates, $0.001$ and $0.03$, by $\{0.4, 0.6, 0.8, 1.0, 1.2, 1.4, 1.6\}$ while holding all other settings fixed. For the gradient-direction analysis, we replace the reward-derived gradients in \model with Gaussian random directions at each optimization step, resulting in a random-walk variant without reward guidance. We report Boxed accuracy averaged over GPQA-Diamond, GSM8K, and MATH-500 using LLaMA-3.2-3B-Instruct, LLaMA-3.1-8B-Instruct, and Qwen3-4B-Instruct.

\textbf{Robustness to Learning Rates.}
As shown in \Cref{fig:learning_rate_robustness}, \model maintains strong performance across all seven learning-rate settings. Its accuracy varies only from \textbf{51.4\% to 53.8\%}, compared with 47.6\% to 51.8\% for \textsc{LatentSeek}. It also achieves both a higher average accuracy (\textbf{52.6\%} versus 49.3\%) and a substantially lower standard deviation (\textbf{0.82} versus 1.53). These results indicate that direct interaction with selected-layer latent states provides a more stable optimization interface that is less sensitive to step-size selection.

\textbf{Robustness to Optimization Directions.}
As shown in \Cref{fig:random_vs_latentseek}, even the random-walk variant of \model achieves an average accuracy of \textbf{60.6\%}, slightly exceeding the 60.3\% of reward-guided \textsc{LatentSeek}. Thus, useful reasoning trajectories can be discovered through direct exploration of the selected-layer latent space even without an explicitly optimized direction. This comparison separates two sources of improvement: the direct latent interaction introduced by \model makes the optimization space inherently more accessible and robust, while reward guidance determines how effectively that space is explored. 

\subsection{Latent Dynamics: Token-Level Gradient Attribution Analysis}
\label{sec:token_gradient_attribution}

We conduct the analysis using LLaMA-3.2-3B-Instruct on GPQA-Diamond, GSM8K, and MATH-500 with the Boxed answer format. Gradients are computed from the generated trajectory at the final latent-optimization step. For each continuation token, we define its gradient strength as the $L_2$ norm of its gradient with respect to all optimized latent states. We then average the gradient strength over tokens assigned to the same category and across evaluation examples. Tokens are automatically classified using the rule-based categories summarized in \Cref{tab:gradient_token_categories}.

\begin{table}[t]
\centering
\fontsize{9pt}{10.5pt}\selectfont

\begin{tabular}{ll}
\toprule
Token category & Representative tokens \\
\midrule
Formatting
& \texttt{.}, \texttt{,}, \texttt{\#}, \texttt{-} \\
\addlinespace

Reasoning Connector
& \texttt{because}, \texttt{therefore}, \texttt{then}, \texttt{however} \\
\addlinespace

Content Explanation
& \texttt{compute}, \texttt{equation}, \texttt{value}, \texttt{number} \\
\addlinespace

Answer Marker
& \texttt{\textbackslash boxed}, \texttt{\#\#\#\#}, \texttt{final}, \texttt{answer} \\
\addlinespace

Answer Content
& \texttt{42}, \texttt{C}, \texttt{3.14}, \texttt{7/8} \\
\bottomrule
\end{tabular}
\caption{Continuation-token categories used in the gradient attribution analysis.}
\label{tab:gradient_token_categories}
\end{table}

\begin{figure*}[t]
    \centering
    \includegraphics[width=\textwidth]{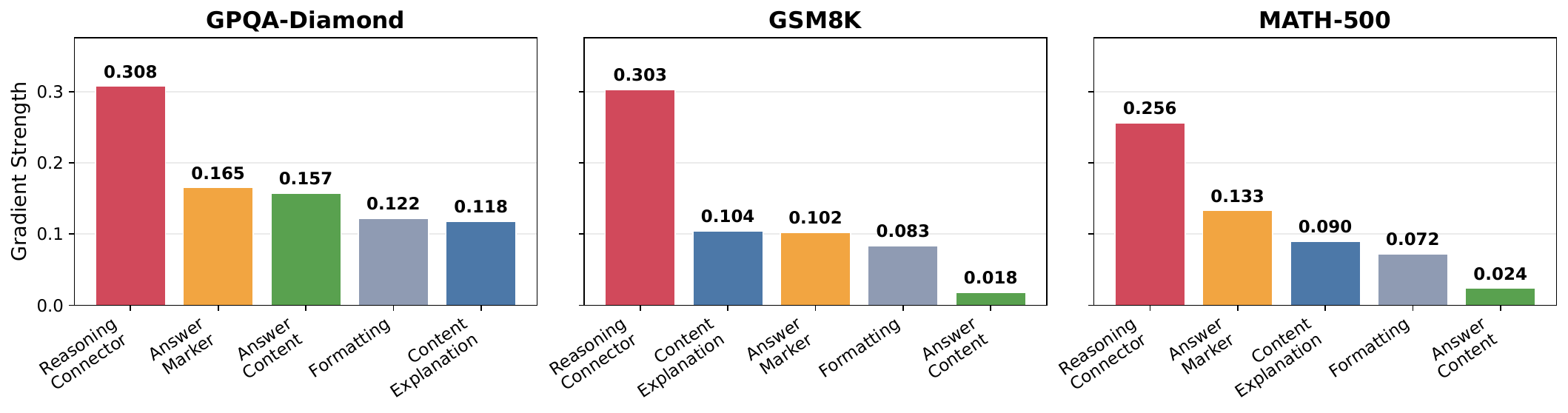}
    \caption{Gradient strength across continuation-token categories on three reasoning benchmarks.}
    \label{fig:token_gradient_attribution}
\end{figure*}

\textbf{Reasoning connectors receive the strongest gradients.}
As shown in \Cref{fig:token_gradient_attribution}, reasoning connectors consistently exhibit the highest gradient strength across all three benchmarks. Tokens such as \emph{because}, \emph{therefore}, and \emph{then} mark transitions between successive reasoning steps. Their consistently high first-order sensitivity indicates that optimized latent states primarily influence how the model connects and advances its reasoning process, rather than affecting continuation tokens uniformly.

\section{Further Analysis}
In this section, we perform deeper analysis on \model.
\subsection{Ablation Study}
\label{sec:ablation_study}

We conduct the ablation study across five backbones and three benchmarks using the Boxed answer format. We compare the full \model with three variants: \emph{w/o Gradient} replaces the reward-derived gradients with Gaussian random directions, \emph{w/o Latent Update} retains the inserted prefix without updating it, and \emph{w/o Inserted Prefix} reduces the method to standard CoT. The results are presented in \Cref{tab:ablation_study}.


\begin{table*}[t]
\centering
\resizebox{\linewidth}{!}{

\newcommand{\abscore}[2]{%
  \begin{tabular}[c]{@{}c@{}}
    #1\\[-0.4ex]
    \textcolor{red}{(-#2)}
  \end{tabular}%
}

\newcommand{\abmethod}[1]{%
  \begin{tabular}[c]{@{}l@{}}
    \mbox{#1}
  \end{tabular}%
}

\begin{tabular}{l*{15}{c}}
\toprule

& \multicolumn{3}{c}{LLaMA-3.2-3B}
& \multicolumn{3}{c}{LLaMA-3.1-8B}
& \multicolumn{3}{c}{Qwen2.5-7B}
& \multicolumn{3}{c}{Qwen2.5-14B}
& \multicolumn{3}{c}{Qwen3-4B} \\
\cmidrule(lr){2-4}
\cmidrule(lr){5-7}
\cmidrule(lr){8-10}
\cmidrule(lr){11-13}
\cmidrule(lr){14-16}

Method
& G & S & M
& G & S & M
& G & S & M
& G & S & M
& G & S & M \\
\midrule

\model
& \textbf{30.3} & \textbf{82.5} & \textbf{53.6}
& \textbf{31.3} & \textbf{86.2} & \textbf{57.4}
& \textbf{38.4} & \textbf{91.8} & \textbf{77.2}
& \textbf{42.4} & \textbf{93.3} & \textbf{80.2}
& \textbf{52.5} & \textbf{90.0} & \textbf{91.8} \\
\addlinespace[2pt]

\abmethod{w/o Gradient}
& \abscore{26.3}{4.0}
& \abscore{80.7}{1.8}
& \abscore{49.4}{4.2}
& \abscore{25.3}{6.0}
& \abscore{85.4}{0.8}
& \abscore{54.2}{3.2}
& \abscore{37.9}{0.5}
& \abscore{89.5}{2.3}
& \abscore{75.8}{1.4}
& \abscore{\textbf{42.4}}{0.0}
& \abscore{93.0}{0.3}
& \abscore{78.8}{1.4}
& \abscore{48.0}{4.5}
& \abscore{89.6}{0.4}
& \abscore{86.2}{5.6} \\
\addlinespace[2pt]

\abmethod{w/o Latent Update and the above}
& \abscore{23.2}{7.1}
& \abscore{78.5}{4.0}
& \abscore{47.6}{6.0}
& \abscore{20.2}{11.1}
& \abscore{82.6}{3.6}
& \abscore{49.0}{8.4}
& \abscore{36.4}{2.0}
& \abscore{89.2}{2.6}
& \abscore{72.2}{5.0}
& \abscore{41.4}{1.0}
& \abscore{92.3}{1.0}
& \abscore{78.4}{1.8}
& \abscore{43.4}{9.1}
& \abscore{89.3}{0.7}
& \abscore{86.8}{5.0} \\
\addlinespace[2pt]

\abmethod{w/o Inserted Prefix and the above}
& \abscore{17.7}{12.6}
& \abscore{75.4}{7.1}
& \abscore{40.4}{13.2}
& \abscore{20.7}{10.6}
& \abscore{81.7}{4.5}
& \abscore{50.6}{6.8}
& \abscore{31.3}{7.1}
& \abscore{89.2}{2.6}
& \abscore{73.0}{4.2}
& \abscore{40.4}{2.0}
& \abscore{92.4}{0.9}
& \abscore{79.2}{1.0}
& \abscore{40.9}{11.6}
& \abscore{88.1}{1.9}
& \abscore{87.2}{4.6} \\

\bottomrule
\end{tabular}
}
\caption{
Ablation results across five instruction-tuned backbones.
We report accuracy (\%). Within each backbone, G, S, and M denote
GPQA-Diamond, GSM8K, and MATH-500, respectively.
}
\label{tab:ablation_study}
\end{table*}

\textbf{The inserted prefix alone is insufficient.}
Averaged over all 15 backbone--benchmark settings, inserting a fixed prefix improves accuracy from \textbf{60.5\%} to \textbf{62.0\%}, but degrades performance in six settings and ties CoT in one. In contrast, the full \model reaches \textbf{66.6\%} and outperforms the fixed-prefix variant in all 15 settings by an average of \textbf{4.6 points}. The improvement therefore cannot be explained solely by the additional textual reasoning cue.

\textbf{Reward guidance provides the decisive gain.}
Random Optimization outperforms the fixed-prefix variant in \textbf{14 of 15 settings}, indicating that modifying selected-layer latent states can expose useful alternative reasoning trajectories. Replacing random directions with reward-derived gradients provides a further average improvement of \textbf{2.4 points}: \model outperforms random optimization in 14 settings, ties it in one, and is best or tied-best in every column. These results indicate that selected-layer optimization and reward-guided credit assignment provide complementary benefits.

\subsection{Analysis of Optimized Layers}
\label{sec:layer_analysis}
\begin{figure*}[t]
    \centering
    \includegraphics[width=\textwidth]{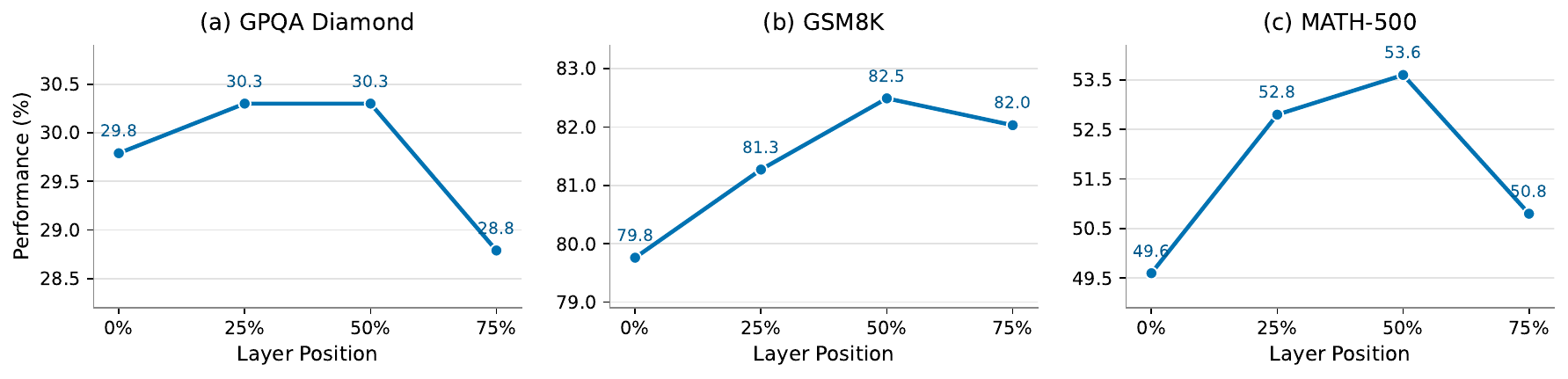}
    \caption{Sensitivity of \model\ to the optimized layer on LLaMA-3.2-3B-Instruct.
    Performance is reported on (a) GPQA-Diamond, (b) GSM8K, and (c) MATH-500.
    The horizontal axis indicates the normalized depth at which the latent is optimized.}
    \label{fig:layer_sensitivity}
\end{figure*}
We investigate how the layer at which the latent prefix is optimized affects the performance of \model.
Using LLaMA-3.2-3B-Instruct as the backbone, we place the optimizable prefix at four normalized depths.

As shown in \Cref{fig:layer_sensitivity}, \textbf{optimizing the latent prefix at an intermediate layer generally outperforms optimization in the embedding space}. On GPQA-Diamond, the 25\% and 50\% placements both achieve the highest accuracy of $30.3\%$, compared with $29.8\%$ at the embedding level. The benefit is most pronounced on MATH-500, where moving the latent toward the middle of the network yields the largest improvement. \looseness=-1

The optimal layer is therefore task-dependent, but the strongest results consistently occur between 25\% and 50\% of the network depth.
Moving the optimization point further to 75\% does not provide an additional gain and can be detrimental, most notably on GPQA-Diamond.
These results are consistent with a favorable trade-off at early-to-middle depths: \textbf{the hidden states are contextualized enough to provide informative optimization signals while retaining sufficient downstream Transformer computation to propagate and refine the effect of the optimized latent}.

\subsection{Case Study}
\label{sec:case_study}

To qualitatively investigate how test-time latent optimization affects
model reasoning, we present a representative example in
Table~\ref{tab:case_study}. We compare the outputs before and after
optimization by LatentSeek and \model.

\begin{table}[t]
\centering

\setlength{\fboxsep}{6pt}

\fbox{%
\begin{minipage}{\dimexpr\columnwidth-2\fboxsep-2\fboxrule\relax}
\small
\setlength{\tabcolsep}{4pt}
\renewcommand{\arraystretch}{1.15}

\begin{tabularx}{\linewidth}{@{}p{0.19\linewidth}X@{}}

\multicolumn{2}{@{}p{\linewidth}@{}}{%
\textbf{Question.}
Find the unique \(\textbf{odd}\) integer \(t\) such that \(0<t<23\) and
\(t+2\) is the inverse of \(t\) modulo \(23\).
}
\\

\midrule

\multicolumn{2}{@{}l@{}}{%
\textsc{LatentSeek}
}
\\[0.35em]

\textbf{Original}
&
\textit{%
We seek the unique integer \(t\) such that \(0<t<23\) and \(t+2\) is the
multiplicative inverse of [...] The final answer is
\textbackslash{}boxed\{19\}.%
}
\\

\addlinespace[0.3em]

\textbf{Optimized}
&
\textit{%
We solve unique integer \(t\), \textcolor{darkred}{that that} \(0<t<23\) and \(t+2\) is the
inverse of [...] modulo 23 is \textbackslash{}boxed\{20\}.%
}
\\

\midrule

\multicolumn{2}{@{}l@{}}{%
\textsc{GradCuit} (Ours)
}
\\[0.35em]

\textbf{Original}
&
\textit{%
We are given that \(t+2\) is the inverse of \(t\) modulo 23. [...]
\(t\) modulo 23 is \textbackslash{}boxed\{20\}.%
}
\\

\addlinespace[0.3em]

\textbf{Optimized}
&
\textit{%
We are given that \(t+2\) is the inverse of \(t\) modulo 23. [...]
The answer is \(t=\textbackslash{}boxed\{17\}\).%
}
\\

\end{tabularx}
\end{minipage}%
}

\caption{%
A representative case comparing the outputs before and after latent
optimization. All model outputs are displayed verbatim, and the
ground-truth answer is \(17\).%
}
\label{tab:case_study}

\end{table}

As shown in Table~\ref{tab:case_study}, LatentSeek changes the
decoded token sequence during optimization but still produces an incorrect
answer and introduces malformed text such as ``that that.'' In contrast,
GradCuit directly updates the hidden states without explicitly
modifying the token sequence, thereby avoiding such textual corruption
while correcting the answer from \(20\) to \(17\). This case further illustrates
the greater robustness of GradCuit.

\section{Related Work}
\label{sec:related_works}

\paragraph{Test-Time Optimization.}
A broad set of methods improve language-model outputs \emph{at inference time} by allocating additional computation.
Discrete approaches include repeated sampling and aggregation, such as self-consistency
\citep{wang2022self}, as well as recitation-augmented generation \citep{sun2023recitation} and
explicit trajectory search \citep{hao2022reasoning}. Continuous inference-time steering includes
PPLM, which updates internal activations using gradients from a differentiable attribute model while
keeping the base language model frozen \citep{dathathri2019plug}. Amulet instead formulates each
token-decoding step as an online optimization problem for test-time preference adaptation
\citep{zhang2025amuletrealignmenttesttime}. CTRL is related to controllable generation, but its
control-code conditioning is learned during pre-training rather than optimized at test time
\citep{keskar2019ctrl}. Test-Time Training (TTT) updates model parameters or fast model state online
using self-supervised objectives \citep{sun2020test,sun2024learning,hardt2023test}. In contrast,
\model keeps all model parameters fixed and optimizes only a small set of instance-specific states
inserted at a selected Transformer layer.

\paragraph{Reinforcement Learning for Language Models.}
Reinforcement learning from human feedback is exemplified by InstructGPT
\citep{ouyang2022training}, whereas Constitutional AI uses reinforcement learning from AI feedback
\citep{bai2022constitutional}. Policy-gradient methods such as PPO are widely used in this setting
\citep{schulman2017proximal}. More recent alignment objectives and algorithms include Direct
Preference Optimization \citep{rafailov2023direct} and Statistical Rejection Sampling
\citep{liu2023statistical}. Group Relative Policy Optimization was introduced in DeepSeekMath
\citep{shao2024deepseekmath} and subsequently used at scale in DeepSeek-R1
\citep{guo2025deepseek_reasoning}.
In parallel, reward modeling has evolved from human preference models to automated and language-model-driven reward design \citep{kwon2023reward}, as well as multi-agent verification frameworks \citep{lifshitz2025multi}. 
A related variational perspective is explored by \citet{chen2024languagemodelshiddenreasoners}, which improves latent trajectory fitting by updating model parameters. 
Compared with these training-time approaches, our method operates entirely at test time: it does not update the policy model, but instead uses gradients of token-level objectives to refine instance-specific latent states.

\paragraph{Latent Reasoning.}
Prompting-based reasoning methods such as chain-of-thought prompting \citep{wei2022chain,kojima2022large,zhou2022least} elicit intermediate reasoning steps in text. 
Compute-adaptive strategies further adjust inference effort based on task complexity \citep{snell2025scaling,misaki2025wider}. 
Motivated by the limitations of explicit textual traces, latent chain-of-thought methods replace or compress reasoning traces into continuous representations \citep{hao2024training,shen2025codi,cheng2024compressed,deng2024explicit}. 
Prompt tuning and soft/prefix prompting \citep{lester2021power,liu2024gpt,li2021prefix,liu2021p} also inject trainable vectors, but they typically require supervised data and training-time backpropagation through the model. 
Our work is closest to test-time latent-optimization methods. LatentSeek optimizes output-side latent representations that are decoded into token sequences for reward evaluation \citep{li2026seekdarkreasoningtesttime}. LTPO optimizes input-level latent thought vectors using a confidence reward and perturbation-based policy gradients \citep{ye2025thinkingflytesttimereasoning}. MILR searches intermediate text--image output representations for multimodal generation \citep{mi2025milrimprovingmultimodalimage}, while DMLR refines latent think tokens using confidence-guided policy gradients and dynamic visual injection \citep{liu2025reasoningminddynamicmultimodal}. GradCuit instead inserts optimizable states at a selected Transformer layer and directly differentiates continuation-token log-probabilities with respect to these states through the remaining Transformer computation.\looseness=-1

\section{Conclusion}

We study the credit-assignment problem in test-time latent reasoning: how to propagate sequence-level rewards to the instance-specific latent states that shape a reasoning trajectory. We introduce \model, which realizes \emph{Credit-Assigned Gradient Flow} by inserting optimizable latent states at a selected Transformer layer. This construction makes every continuation-token log-probability differentiable with respect to every latent state, enabling direct reward-weighted optimization while keeping the base model frozen.

Across five backbones and three reasoning benchmarks, \model achieves the highest average accuracy among all evaluated methods while exhibiting substantially lower sensitivity to learning-rate selection than LatentSeek. Even when reward-derived gradients are replaced with random directions, \model remains competitive with LatentSeek, indicating that direct interaction within a selected-layer latent space provides a robust optimization interface. Token-level gradient attribution shows that latent influence concentrates on reasoning-connector tokens, while ablation studies establish the complementary roles of the inserted prefix, latent-state optimization, and reward guidance. Additional analyses identify early-to-middle Transformer layers as the most effective optimization space and illustrate, through a representative case, that direct hidden-state updates can correct an erroneous answer without introducing malformed text. Together, these findings demonstrate that \model enables \textbf{more robust and interpretable} test-time latent reasoning.

\clearpage

{
\small

\bibliographystyle{unsrtnat}
\bibliography{ref}
}
\newpage

\newcolumntype{Y}{>{\raggedright\arraybackslash}X}

\appendix

\section{Implementation Details}
\label{app:implementation}

\subsection{Details of GradCuit}
\label{app:details_of_gradcuit}

\paragraph{Latent Construction and Deployment.}
For each problem, we first render the system and user messages using the native Hugging Face chat template of the corresponding backbone.
We then append the fixed textual prefix, ``Let's think about this problem and solve it step by step.'', after the rendered prompt and before the generated continuation.
The number of inserted latent states, denoted by $N$, is therefore determined by the number of tokens into which each backbone tokenizer decomposes this prefix.
The optimized layer $l$ is a method hyperparameter that determines the decoder-block input at which the prefix representations are treated as optimizable latent states.
For a model with $M$ decoder blocks, we set $l=\lfloor M/2 \rfloor$ in the main experiments to obtain a consistent representative configuration across backbones.
We additionally vary $l$ in the layer-position analysis to examine how the location of the latent optimization space affects optimization performance.
We initially pass the prompt and textual prefix through the first $l$ Transformer blocks and extract the hidden representations corresponding to the prefix span, denoted by
$\boldsymbol{z}_{0}^{(l)}\in \mathbb{R}^{N\times d}$.
Rather than repeatedly encoding the textual prefix, we introduce a trainable offset $\Delta\boldsymbol{z}^{(l)}$, initialized to zero, and use
\[
\widetilde{\boldsymbol{z}}^{(l)}
=
\boldsymbol{z}_{0}^{(l)}
+
\Delta\boldsymbol{z}^{(l)}
\]
as the prefix representation throughout optimization.
At the input to the selected decoder block, a forward pre-hook replaces the original prefix span with $\widetilde{\boldsymbol{z}}^{(l)}$.
All prompt and continuation representations retain their original positions, and the standard causal attention mask is used without modification.
Consequently, continuation tokens can attend to all inserted latent states while the model architecture and parameters remain unchanged.
All model parameters are frozen, and gradients are computed only with respect to $\Delta\boldsymbol{z}^{(l)}$.

\paragraph{Test-Time Optimization.}
Before optimization, we greedily generate an initial continuation using $\boldsymbol{z}_{0}^{(l)}$ and evaluate its final answer with the self-reward verifier described below.
At each optimization step, the currently generated token sequence $\boldsymbol{x}=(x_1,\ldots,x_T)$ is treated as fixed and passed through the model using teacher forcing.
We disable the key--value cache during this gradient computation and collect the log-probability assigned to every continuation token.
A single trajectory is used at each step to estimate the sequence-level objective, and the implementation minimizes
\begin{equation}
\label{eq:gradcuit_objective}
\mathcal{L}_{\mathrm{opt}}
=
-R(\boldsymbol{x},\boldsymbol{c})
\sum_{t=1}^{T}
\log\pi\left(
x_t
\mid
\boldsymbol{x}_{<t},
\widetilde{\boldsymbol{z}}^{(l)},
\boldsymbol{c}
\right).
\end{equation}
The token log-probabilities are summed without length normalization.
The generated token IDs and their input embeddings are detached from the computational graph, as is the scalar reward, so the gradient propagates only from the continuation-token log-probabilities through the remaining Transformer blocks to the latent offset.
We update $\Delta\boldsymbol{z}^{(l)}$ using Adam with a learning rate of $10^{-3}$, the default coefficients $\beta_1=0.9$ and $\beta_2=0.999$, and $\epsilon=10^{-8}$.
No weight decay, gradient clipping, latent-norm constraint, or additional regularization is applied.
After every update, we greedily regenerate the complete continuation using the updated latent states and obtain a new reward.
Optimization terminates immediately once the verifier accepts the final answer or after at most ten latent updates.
If no answer is accepted within this budget, the continuation produced after the final update is returned.

\begin{algorithm}[t]
\caption{GradCuit inference for one problem instance.}
\label{alg:gradcuit_inference}
\begin{algorithmic}[1]
\Require Problem context $\boldsymbol{c}$ and frozen model $\pi$
\Require Prefix $s$, optimized layer $l$, learning rate $\eta$, and update budget $K$
\State Render the model-specific chat prompt and append $s$
\State Extract the prefix states $\boldsymbol{z}_{0}^{(l)}$ at the input to decoder block $l$
\State Initialize $\Delta\boldsymbol{z}^{(l)} \gets \boldsymbol{0}$
\State $\widetilde{\boldsymbol{z}}^{(l)} \gets \boldsymbol{z}_{0}^{(l)} + \Delta\boldsymbol{z}^{(l)}$
\State Generate $\boldsymbol{x}$ greedily using $\widetilde{\boldsymbol{z}}^{(l)}$
\State Obtain $R(\boldsymbol{x},\boldsymbol{c})$ from the verifier
\For{$k=1,\ldots,K$}
    \If{$R(\boldsymbol{x},\boldsymbol{c})=0$}
        \State \Return $\boldsymbol{x}$
    \EndIf
    \State Recompute continuation-token log-probabilities with teacher forcing
    \State Compute $\mathcal{L}_{\mathrm{opt}}$ using Eq.~\eqref{eq:gradcuit_objective}
    \State Update $\Delta\boldsymbol{z}^{(l)}$ by one Adam step with learning rate $\eta$
    \State $\widetilde{\boldsymbol{z}}^{(l)} \gets \boldsymbol{z}_{0}^{(l)} + \Delta\boldsymbol{z}^{(l)}$
    \State Regenerate $\boldsymbol{x}$ greedily using $\widetilde{\boldsymbol{z}}^{(l)}$
    \State Re-evaluate $R(\boldsymbol{x},\boldsymbol{c})$
\EndFor
\State \Return $\boldsymbol{x}$
\end{algorithmic}
\end{algorithm}

\paragraph{Self-Reward Verifier.}
Across all experiments that require a reward or verification signal, \textbf{the evaluated LLM itself} serves as the self-verifier and supplies the self-reward; no separate reward model is introduced.
For each backbone, GradCuit, Self-Reflection, Self-Scored BoN, and LatentSeek all invoke that same backbone using an identical verifier prompt, answer-extraction procedure, and decision rule.
The verifier receives only the original question and the extracted final answer.
It evaluates final-answer correctness without scoring the completeness or quality of the generated reasoning.
Its binary verdict is mapped to the discrete reward
\[
R(\boldsymbol{x},\boldsymbol{c})
=
\begin{cases}
0,  & \text{correct},\\
-1, & \text{otherwise}.
\end{cases}
\]
An output from which no valid final answer can be extracted is assigned a reward of $-1$.
This unified self-reward protocol ensures that performance differences among GradCuit, Self-Reflection, Self-Scored BoN, and LatentSeek do not arise from different reward mechanisms.
The complete verifier prompt is:

\begin{quote}
\raggedright
\ttfamily
You are a critical verifier for mathematical questions.\par
You will be given the original question and one final answer.\par
Decide whether that answer is correct for the question.\par
\medskip
QUESTION:\par
\{question\}\par
\medskip
FINAL ANSWER:\par
\{extracted\_answer\}\par
\medskip
INSTRUCTIONS:\par
1. Verify only the final answer. Do not evaluate any missing reasoning steps.\par
2. Do not solve the problem independently from scratch unless a tiny auxiliary calculation is strictly necessary for verification.\par
3. Prefer reverse verification methods such as substitution, plugging the answer back into the original conditions, checking algebraic consistency, checking boundary cases, or other direct validation targeted at the proposed answer.\par
4. Accept mathematically equivalent forms when they represent the same final answer.\par
5. If the final answer is correct, the verdict is True.\par
6. If the final answer is incorrect, the verdict is False.
\end{quote}

\paragraph{Generation Prompts and Decoding.}
For the Boxed format, the system and user messages are:

\begin{quote}
\raggedright
\ttfamily
\{"role": "system", "content": "Please reason step by step, and put your final answer within \textbackslash boxed\{\}."\}\par
\{"role": "user", "content": q\}
\end{quote}

For the JSON format, the system and user messages are:

\begin{quote}
\raggedright
\ttfamily
\{"role": "system", "content": "Please reason step by step, and put your final answer in a json dict with exactly one key "answer", for example \{"answer": "1.234"\}."\}\par
\{"role": "user", "content": q\}
\end{quote}

GradCuit uses greedy decoding with a batch size of one.
We set the maximum continuation length to 4,096 tokens for Qwen3-4B-Instruct-2507 because it typically produces longer reasoning trajectories, and to 2,048 tokens for all other backbones.
To ensure a fair comparison, these backbone-specific maximum continuation lengths are kept identical across GradCuit and all baselines, while each method retains its required decoding strategy.

\begin{table*}[t]
\centering
\fontsize{9pt}{10.5pt}\selectfont
\caption{Backbone-specific GradCuit configurations used in the main experiments.
Layer indices identify the decoder-block input at which the inserted prefix states are optimized.}
\label{tab:gradcuit_backbone_configuration}
\begin{tabular}{lccc}
\toprule
Backbone
& Decoder Blocks
& Optimized Layer
& Maximum New Tokens \\
\midrule
LLaMA-3.2-3B-Instruct
& 28
& 14
& 2,048 \\
LLaMA-3.1-8B-Instruct
& 32
& 16
& 2,048 \\
Qwen2.5-7B-Instruct
& 28
& 14
& 2,048 \\
Qwen2.5-14B-Instruct
& 48
& 24
& 2,048 \\
Qwen3-4B-Instruct-2507
& 36
& 18
& 4,096 \\
\bottomrule
\end{tabular}
\end{table*}

\paragraph{Random-Direction Variant.}
For the random-direction analysis, the reward-derived update is replaced at every step by an independently sampled Gaussian direction.
Specifically, with
$\boldsymbol{\epsilon}_k\sim\mathcal{N}(\boldsymbol{0},\mathbf{I})$,
we apply
\[
\Delta\boldsymbol{z}_{k+1}^{(l)}
=
\Delta\boldsymbol{z}_{k}^{(l)}
+
\eta
\frac{\boldsymbol{\epsilon}_k}
{\lVert\boldsymbol{\epsilon}_k\rVert_2}.
\]
The sampled direction is globally $L_2$-normalized and then scaled by the learning rate.
This variant does not use Adam or match the random-vector norm to the reward-derived gradient norm.
The verifier is retained only for the common stopping rule.
All remaining settings, including the inserted prefix, optimized layer, learning rate, generation budget, and decoding strategy, are identical to those of GradCuit.

\paragraph{Numerical and Randomness Settings.}
Backbone inference and latent states use bfloat16 precision.
The main experiments use a fixed random seed of 42 and report one run for each backbone--benchmark--format configuration.
For the layer-position analysis with LLaMA-3.2-3B-Instruct, the 0\%, 25\%, 50\%, and 75\% positions correspond to the word-embedding space and decoder-block input indices 7, 14, and 21, respectively.

\subsection{Details of Other Baselines}
\label{app:details_of_baselines}

\paragraph{Shared Protocol.}
All baselines use the native Hugging Face chat template of the corresponding backbone and the same Boxed and JSON system prompts used by GradCuit.
Except for interactions required by each method, no additional instructions are added.
The backbone-specific maximum generation lengths are also kept identical across methods: 4,096 tokens for Qwen3-4B-Instruct-2507 and 2,048 tokens for all other backbones.
For methods requiring answer verification, the currently evaluated backbone itself serves as the self-verifier.
The verifier uses the same prompt and binary reward rule described above, with greedy decoding and a maximum generation length of 8,192 tokens.
All experiments use a fixed random seed of 42.

\paragraph{Chain-of-Thought.}
The Chain-of-Thought (CoT) baseline directly generates one complete response from the corresponding Boxed or JSON prompt using greedy decoding.
It does not include the fixed prefix introduced by GradCuit and performs no subsequent verification, reflection, sampling, or test-time optimization.
The single generated response is directly used as the final output.

\paragraph{One-Pass Self-Reflection.}
The initial response of Self-Reflection is generated using exactly the same procedure as CoT.
The self-verifier then evaluates the extracted final answer.
If the answer is accepted, the initial response is returned without further generation.
Otherwise, we continue the existing multi-turn conversation by retaining the initial response as an assistant message and appending one user turn that asks the backbone to reconsider and revise its answer.
The verifier supplies no textual critique or auxiliary reasoning to the backbone.
The backbone generates one revised response using greedy decoding, which is returned as the final output.
No additional reflection round is performed.

\paragraph{Self-Consistency.}
Self-Consistency generates five candidate responses for each problem using stochastic decoding with $\texttt{temperature}=1.0$.
We explicitly enable sampling and leave all other Hugging Face generation parameters at their default values.
A final answer is extracted from each candidate according to the corresponding Boxed or JSON format.
Candidates from which no valid answer can be extracted are discarded.
The remaining extracted answers are compared using strict string matching, without invoking the verifier or performing additional mathematical-equivalence checking.
The answer occurring most frequently is returned as the final prediction.
If multiple answers receive the same highest number of votes, the one appearing first in generation order is selected.

\paragraph{Self-Scored Best-of-N.}
Self-Scored Best-of-N (BoN) generates 5 candidate responses using the same stochastic decoding configuration as Self-Consistency.
All 5 candidates are generated and subsequently evaluated by the same self-verifier used for GradCuit.
The verifier verdict for each candidate is mapped to the shared discrete reward, with accepted and rejected candidates receiving rewards of $0$ and $-1$, respectively.
The candidate with the highest reward is selected, and its complete generated response is returned as the final output.
When multiple candidates receive the same highest reward, the final candidate in generation order is selected.

\paragraph{LatentSeek.}
We use the official implementation of LatentSeek \citep{li2026seekdarkreasoningtesttime} and retain its recommended method-specific configuration.
LatentSeek uses the same backbone, Boxed or JSON input prompt, answer parser, self-verifier, and backbone-specific maximum generation length as GradCuit.
Its latent variables are optimized using Adam with a learning rate of $0.03$ for at most ten optimization steps.
All remaining method-specific implementation details follow the official LatentSeek implementation.

\subsection{Hardware and Software.}
All experiments were conducted on a server equipped with eight NVIDIA L40 GPUs.
The software environment used Python 3.10, PyTorch 2.1.0, and CUDA 13.2.

\section{Compute Budget and Efficiency}
\label{app:compute_budget}

\begin{table*}[t]
\centering

\fontsize{8pt}{10.5pt}\selectfont
\begin{tabular}{lcccccccccccc}
\toprule
Method
& \multicolumn{2}{c}{LLaMA3.2-3B}
& \multicolumn{2}{c}{LLaMA3.1-8B}
& \multicolumn{2}{c}{Qwen2.5-7B}
& \multicolumn{2}{c}{Qwen2.5-14B}
& \multicolumn{2}{c}{Qwen3-4B}
& \multicolumn{2}{c}{Avg.} \\
\cmidrule(lr){1-1}
\cmidrule(lr){2-3}
\cmidrule(lr){4-5}
\cmidrule(lr){6-7}
\cmidrule(lr){8-9}
\cmidrule(lr){10-11}
\cmidrule(lr){12-13}
Prompt Type
& Boxed & JSON
& Boxed & JSON
& Boxed & JSON
& Boxed & JSON
& Boxed & JSON
& Boxed & JSON \\
\midrule

\multicolumn{13}{c}{GPQA-Diamond} \\
\midrule

\textsc{GradCuit} (Ours)
& 2.64 & 2.68
& 3.63 & 3.25
& 1.79 & 1.85
& 2.49 & 1.98
& 4.98 & 4.52
& 3.11 & 2.86 \\


Self-Consistency
& 5 & 5
& 5 & 5
& 5 & 5
& 5 & 5
& 5 & 5
& 5 & 5 \\

Self-Scored BoN
& 5 & 5
& 5 & 5
& 5 & 5
& 5 & 5
& 5 & 5
& 5 & 5 \\

\midrule
\multicolumn{13}{c}{GSM8K} \\
\midrule

\textsc{GradCuit} (Ours)
& 2.34 & 2.02
& 1.94 & 1.83
& 1.73 & 1.73
& 1.77 & 1.87
& 1.36 & 1.32
& 1.83 & 1.75 \\


Self-Consistency
& 5 & 5
& 5 & 5
& 5 & 5
& 5 & 5
& 5 & 5
& 5 & 5 \\

Self-Scored BoN
& 5 & 5
& 5 & 5
& 5 & 5
& 5 & 5
& 5 & 5
& 5 & 5 \\

\midrule
\multicolumn{13}{c}{MATH-500} \\
\midrule

\textsc{GradCuit} (Ours)
& 2.75 & 3.59
& 3.46 & 3.35
& 2.18 & 2.37
& 1.99 & 1.39
& 1.62 & 2.05
& 2.40 & 2.55 \\


Self-Consistency
& 5 & 5
& 5 & 5
& 5 & 5
& 5 & 5
& 5 & 5
& 5 & 5 \\

Self-Scored BoN
& 5 & 5
& 5 & 5
& 5 & 5
& 5 & 5
& 5 & 5
& 5 & 5 \\

\bottomrule
\end{tabular}

\caption{
Average optimization iterations and repeated-sampling budgets in the main experiments.
For \textsc{GradCuit}, the values include the original unoptimized generation as the first round, followed by any subsequent optimized generations before termination.
For Self-Consistency and Self-Scored BoN, the values report the fixed number of sampled responses.
}
\label{tab:compute_iterations}
\end{table*}
\begin{table}[h!]
\centering

\setlength{\fboxsep}{6pt}

\fbox{%
\begin{minipage}{\dimexpr\columnwidth-2\fboxsep-2\fboxrule\relax}
\small
\setlength{\tabcolsep}{3pt}
\renewcommand{\arraystretch}{1.15}

\begin{tabularx}{\linewidth}{@{}p{0.225\linewidth}Y@{}}

\multicolumn{2}{@{}p{\linewidth}@{}}{%
\textbf{Question.}
Which sequence of reactions would lead to the high-yield synthesis of
1-(3-bromo-5-nitrophenyl)ethan-1-one from benzene?

\textbf{Choices:}
(A) i) HNO$_3$/H$_2$SO$_4$; ii) Fe/HCl; iii) NaNO$_2$/HCl;
iv) H$_3$PO$_2$; v) Br$_2$/FeBr$_3$; vi) CH$_3$COCl/AlCl$_3$;
vii) HNO$_3$/H$_2$SO$_4$

(B) i) CH$_3$COCl/AlCl$_3$; ii) Br$_2$/FeBr$_3$;
iii) HNO$_3$/H$_2$SO$_4$; iv) Fe/HCl; v) HNO$_3$/H$_2$SO$_4$;
vi) NaNO$_2$/HCl; vii) H$_3$PO$_2$

(C) i) HNO$_3$/H$_2$SO$_4$; ii) Fe/HCl; iii) CH$_3$COCl/AlCl$_3$;
iv) Br$_2$/FeBr$_3$; v) HNO$_3$/H$_2$SO$_4$;
vi) NaNO$_2$/HCl; vii) H$_3$PO$_2$

(D) i) Br$_2$/FeBr$_3$; ii) HNO$_3$/H$_2$SO$_4$;
iii) CH$_3$COCl/AlCl$_3$; iv) HNO$_3$/H$_2$SO$_4$; v) Fe/HCl;
vi) NaNO$_2$/HCl; vii) H$_3$PO$_2$
}
\\

\midrule

\textbf{Original}
&
\textit{%
1. The target compound is 1-(3-bromo-5-nitrophenyl)ethan-1-one, which is
an aromatic ketone with a bromo and a nitro group on the benzene ring.
[...] The closest sequence that matches the above steps is option (D),
but it includes an extra step (v) HNO3/H2SO4, which is not necessary.
Therefore, the correct answer is \textbackslash{}boxed\{D\}.%
}
\\

\addlinespace[0.3em]

\textbf{Optimized}
&
\textit{%
The target compound is 1-(3-bromo-5-nitrophenyl)ethan-1-one. To
synthesize this compound, we need to introduce a nitro group (-NO2)
[...] A, B and D options do not include the correct sequence of
reactions. Therefore, the correct answer is:
\textbackslash{}boxed\{C\}.%
}
\\

\end{tabularx}
\end{minipage}%
}

\caption{%
A case from GPQA-Diamond. The backbone is Llama3.1-8B-Instruct and the ground-truth answer is C.%
}
\label{tab:case_gpqa_synthesis}

\end{table}

\begin{table}[h!]
\centering

\setlength{\fboxsep}{6pt}

\fbox{%
\begin{minipage}{\dimexpr\columnwidth-2\fboxsep-2\fboxrule\relax}
\small
\setlength{\tabcolsep}{3pt}
\renewcommand{\arraystretch}{1.15}

\begin{tabularx}{\linewidth}{@{}p{0.225\linewidth}Y@{}}

\multicolumn{2}{@{}p{\linewidth}@{}}{%
\textbf{Question.}
A merchant must choose between purchasing jewelry worth \(\$5{,}000\)
and electronic gadgets worth \(\$8{,}000\). The jewelry market is
expected to rise by \(2.5\%\), while the electronics market is expected
to rise by \(1.2\%\) within the month. What is the maximum profit the
merchant can obtain?
}
\\

\midrule

\textbf{Original}
&
\textit{%
Step 1: Calculate the future value of the jewelry after a 2.5\% increase.
[...] Profit = \$8,096 - \$8,000 Profit = \$96 Therefore, the merchant
will make a profit of \$96 by choosing the electronic gadgets.
\textbackslash{}boxed\{\$96\}%
}
\\

\addlinespace[0.3em]

\textbf{Optimized}
&
\textit{%
1. The merchant has two options: - Option 1: Purchase jewelry worth
\$5,000. [...] 9. Since \$125 is greater than \$96, the merchant should
choose Option 1 (jewelry) to maximize profit. 10. Therefore, the maximum
profit the merchant can make is \$125.
\textbackslash{}boxed\{\$125\}%
}
\\

\end{tabularx}
\end{minipage}%
}

\caption{%
A case from GSM8K. The backbone is Llama3.1-8B-Instruct and the ground-truth answer is \(\$125\).%
}
\label{tab:case_gsm8k_merchant}

\end{table}

Self-Consistency and Self-Scored BoN use a fixed sampling budget of five responses and generate all five candidates for every problem.
In contrast, \textsc{GradCuit} first generates an unoptimized response and then iteratively updates the latent states until the self-verifier accepts the answer.

Table~\ref{tab:compute_iterations} counts the initial response as the first round and includes all subsequent optimized generations.
Across the 30 backbone--benchmark--format settings, \textsc{GradCuit} averages 2.42 rounds, ranging from 1.32 to 4.98 and remaining below five in every setting.
These results show that its improvements do not rely on more reasoning rounds.

\section{Additional Cases}
\label{app:case_studies}

The main text presents a representative case from MATH-500; here, we provide additional cases from the other two benchmarks, GPQA-Diamond and GSM8K.
In the GPQA-Diamond case shown in Table~\ref{tab:case_gpqa_synthesis}, the original output selects option D despite identifying an inconsistency in its reaction sequence, whereas the optimized output correctly reasons about the order in which the substituents should be introduced and selects option C.
In the GSM8K case shown in Table~\ref{tab:case_gsm8k_merchant}, the original output incorrectly compares the final values of the two investments and selects the electronic gadgets, while the optimized output instead compares their respective profits of \(\$125\) and \(\$96\), yielding the correct maximum profit of \(\$125\).
These cases further illustrate that hidden-state optimization of \textsc{GradCuit} can reorganize the overall reasoning structure and strategy, whereas LatentSeek modifies the early decoded token sequence and may consequently introduce malformed or duplicated text.

\end{document}